\documentclass[11pt,a4paper]{article}
\usepackage[hyperref]{emnlp2020}
\usepackage{times}
\usepackage{latexsym}
\usepackage{booktabs}
\usepackage{multirow}
\usepackage{amsmath}
\usepackage{algorithm}
\usepackage{algorithmic}
\usepackage{graphicx}
\usepackage{subfigure}
\usepackage{makecell}
\usepackage{bm}
\usepackage{url}

\usepackage{microtype}

\aclfinalcopy 

\title{Train No Evil: Selective Masking for Task-Guided Pre-Training}

 \author{Yuxian Gu$^{1,2,3}$, Zhengyan Zhang$^{1,2,3}$, Xiaozhi Wang$^{1,2,3}$, Zhiyuan Liu$^{1,2,3\dagger}$, Maosong Sun$^{1,2,3}$\\
        $^1$Department of Computer Science and Technology, Tsinghua University, Beijing, China  \\
         $^2$Institute for Artificial Intelligence, Tsinghua University, Beijing, China \\ 
         $^3$State Key Lab on Intelligent Technology and Systems, Tsinghua University, Beijing, China\\
         \texttt{\{gu-yx17, zy-z19, wangxz20\}@mails.tsinghua.edu.cn}\\
}

\date{}

\begin{document}
\maketitle
\begin{abstract}
Recently, pre-trained language models mostly follow the pre-train-then-fine-tuning paradigm and have achieved great performance on various downstream tasks. However, since the pre-training stage is typically task-agnostic and the fine-tuning stage usually suffers from insufficient supervised data, the models cannot always well capture the domain-specific and task-specific patterns. In this paper, we propose a three-stage framework by adding a task-guided pre-training stage with selective masking between general pre-training and fine-tuning. 
In this stage, the model is trained by masked language modeling on in-domain unsupervised data to learn domain-specific patterns and we propose a novel selective masking strategy to learn task-specific patterns. Specifically, we design a method to measure the importance of each token in sequences and selectively mask the important tokens.
Experimental results on two sentiment analysis tasks show that our method can achieve comparable or even better performance with less than 50\% of computation cost, which indicates our method is both effective and efficient. The source code of this paper can be obtained from \url{https://github.com/thunlp/SelectiveMasking}.
\end{abstract}

\section{Introduction}

{\let\thefootnote\relax\footnotetext{$^\dagger$ Corresponding author: Z.Liu (liuzy@tsinghua.edu.cn)}}

Pre-trained Language Models (PLMs) have achieved superior performances on various NLP tasks~\cite{baevski2019cloze,joshi2019spanbert,liu2019roberta,yang2019xlnet,Clark2020ELECTRA:} and have attracted wide research interests. Inspired by the success of GPT~\cite{radford2018improving} and BERT~\cite{devlin2018bert}, most PLMs follow the pre-train-then-fine-tuning paradigm, which adopts unsupervised pre-training on large general-domain corpora to learn general language patterns and supervised fine-tuning to adapt to downstream tasks. 

Recently, \citet{gururangan2020dont} shows that learning domain-specific and task-specific patterns during pre-training can be helpful to the models for certain domains and tasks. However, conventional pre-training is aimless with respect to specific downstream tasks, and fine-tuning usually suffers from insufficient supervised data, preventing PLMs from effectively capturing these patterns. 


\begin{figure}[t]
    \centering
    \includegraphics[scale=0.28]{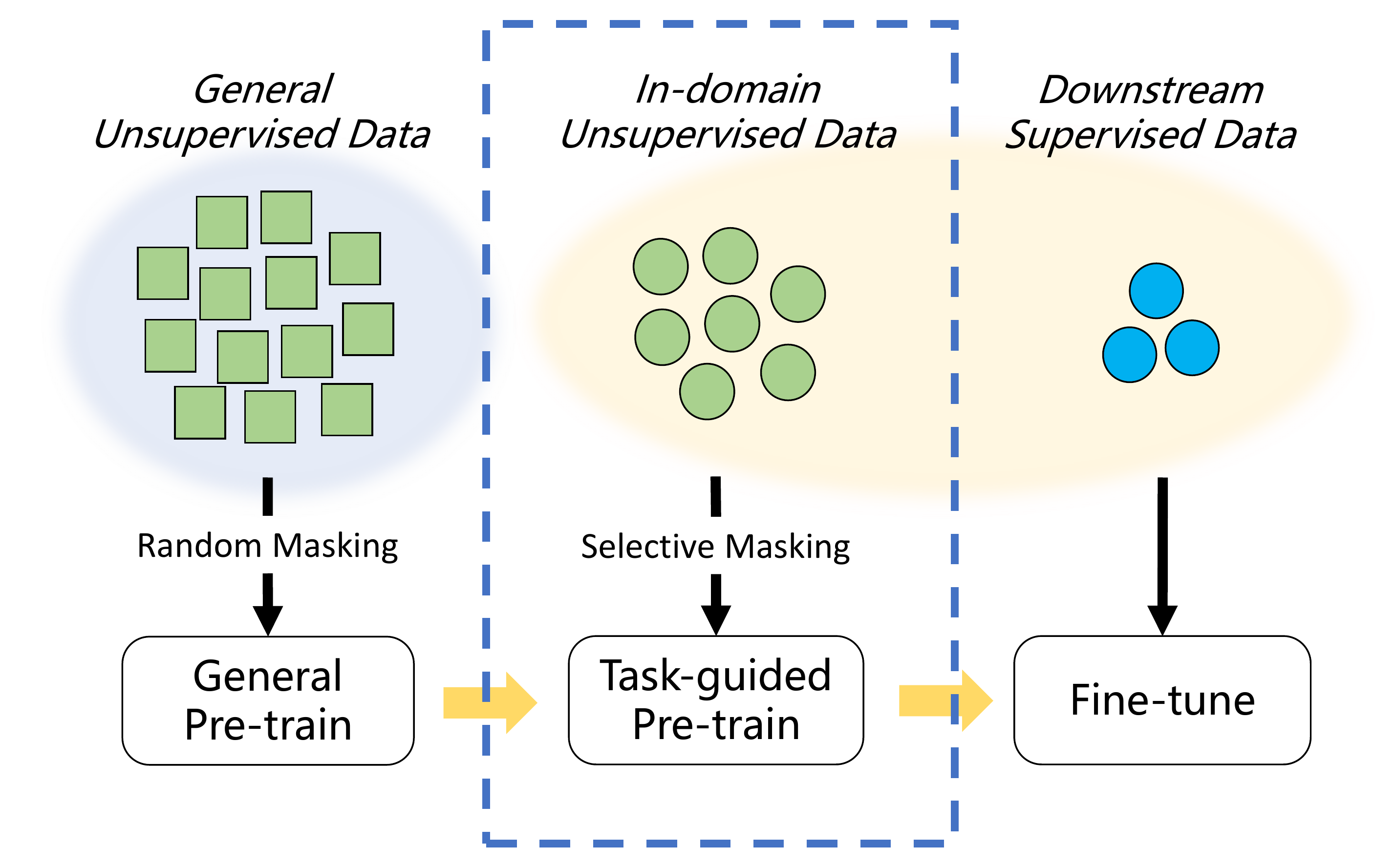}
    \caption{The overall three-stage framework. We add task-guided pre-training between general pre-training and fine-tuning to efficiently and effectively learn the domain-specific and task-specific language patterns.}
    \label{fig:setting}
\end{figure}

To learn \textit{domain-specific language patterns}, some previous works~\cite{Beltagy2019SciBERT,huang2019clinicalbert} pre-train a BERT-like model from scratch using large-scale in-domain data. However, they are computation-intensive and require large-scale in-domain data, which is hard to obtain in many domains. To learn \textit{task-specific 
language patterns}, some previous works~\cite{phang2018sentence} add intermediate supervised pre-training after general pre-training, whose pre-training task is similar to the downstream task but has a larger dataset. However, \citet{wang2019can} shows that this kind of intermediate pre-training often negatively impacts the transferability to downstream tasks.


To better capture domain-specific and task-specific patterns, we propose a three-stage framework by adding a \textbf{task-guided pre-training} stage with selective masking between the general pre-training and fine-tuning. The overall framework is shown in Figure~\ref{fig:setting}. In the stage of task-guided pre-training, the model is trained by masked language modeling (Masked LM)~\cite{devlin2018bert} on mid-scale in-domain unsupervised data, which is constructed by collecting other corpora in the same domain. 
In this way, PLMs can utilize more data to better learn domain-specific language patterns~\cite{alsentzer-etal-2019-publicly,Lee2019BioBERT,sung-etal-2019-pre,xu-etal-2019-doubletransfer, aharoni2020unsupervised}. 
However, the conventional Masked LM randomly masks tokens, which is inefficient to learn task-specific language patterns. Hence, we propose a \textbf{selective masking strategy} for task-guided pre-training, whose main idea is selectively masking the important tokens for downstream tasks.


Intuitively, some tokens are more important than others for a specific task and the important tokens vary among different tasks~\cite{ziser-reichart-2018-pivot, feng-etal-2018-pathologies, rietzler2020adapt}. For instance, in sentiment analysis, sentiment tokens such as ``like'' and ``hate'' are critical for sentiments classification~\cite{ke2019sentilr}. And, in relation extraction, predicates and verbs are typically more significant. If PLMs can selectively mask and predict the important tokens instead of a mass of random tokens, they can effectively learn task-specific language patterns and the computation cost of pre-training can be significantly reduced.

For the selective masking strategy, we propose a simple method to find important tokens for downstream tasks. Specifically, we define a task-specific score for each token and if the score is lower than a certain threshold, we regard the token as important.
However, this method relies on the supervised downstream datasets whose sizes are limited for pre-training. To better utilize mid-scale in-domain unsupervised data as shown in Figure \ref{fig:setting}, we train a neural network on downstream datasets where the important tokens are annotated using the method mentioned above. This neural network can learn the implicit token-selecting rules, which enables us to select tokens without supervision.


We conduct experiments on two sentiment analysis tasks: MR~\cite{pang2005seeing} and SemEval14 task 4~\cite{pontiki-etal-2014-semeval}. Experimental results show that our method is both efficient and effective. Our method can achieve comparable and even better performances than the conventional pre-train-then-fine-tune method with less than 50\% of the overall computation cost.

\section{Methodology}
In this section, we describe task-guided pre-training and selective masking strategy in detail. For convenience, we denote general unsupervised data, in-domain unsupervised data, downstream supervised data as $D_{\mathrm{General}}$, $D_{\mathrm{Domain}}$ and $D_{\mathrm{Task}}$. They generally contain about 1000M words, 10M words, and 10K words respectively.

\subsection{Training Framework}
\label{sec:tf}
As shown in Figure~\ref{fig:setting}, our overall training framework consists of three stages:

\textbf{General pre-training (GenePT)} is identical to the pre-training of BERT~\cite{devlin2018bert}. We randomly mask 15\% tokens of $D_{\mathrm{General}}$ and train the model to reconstruct the original text. 

\textbf{Task-guided pre-training (TaskPT)} trains the model on the mid-scale $D_{\mathrm{Domain}}$ with selective masking to efficiently learn domain-specific and task-specific language patterns. In this stage, we apply a selective masking strategy to focus on masking the important tokens and then train the model to reconstruct the input. The details of selective masking are introduced in Section~\ref{sec:sm}.

\textbf{Fine-tuning} is to adapt the model to the downstream task. This stage is identical to the fine-tuning of the conventional PLMs.

Since TaskPT enables the model to efficiently learn the domain-specific and task-specific patterns, it is unnecessary to fully train the model in the stage of GenePT. Hence, our overall pre-training time cost of the two pre-training stages can be much smaller than those of conventional PLMs.

\subsection{Selective Masking}
\label{sec:sm}
In our TaskPT, we select important tokens of $D_{\mathrm{Task}}$ by their impacts on the classification results. However, this method relies on the supervised labels of $D_{\mathrm{Task}}$. To selectively mask on mid-scale unlabeled in-domain data $D_{\mathrm{Domain}}$, we adopt a neural model to learn the implicit scoring function from the selection results on $D_{\mathrm{Task}}$ and use the model to find important tokens of $D_{\mathrm{Domain}}$.

\subsubsection*{Finding important tokens}\label{heuristic}
We propose a simple method to find important tokens of $D_{\mathrm{Task}}$. Given the $n$-token input sequence $\bm s = (w_1, w_2, \ldots, w_n)$, we use an auxiliary sequence buffer $\bm s'$ to help evaluating these tokens one by one. At time step $0$, $\bm s'$ is initialized to empty. Then, we sequentially add each token $w_i$ to $\bm s'$ and calculate the task-specific score of $w_i$, which is denoted by $\text{S}(w_i)$. If the score is lower than a threshold $\delta$, we regard $w_i$ as an important token. Note that we will remove previous important tokens from $\bm s'$ to make sure the score is not influenced by previous important tokens.

Assume the buffer at the time step $i-1$ is $\bm s_{i-1}'$. We define the token $w_i$'s score as the difference of classification confidences between the original input sequence $\bm s$ and the buffer after adding $w_i$, which is denoted by $\bm s'_{i-1}w_i$:
\begin{equation}
\setlength{\abovedisplayskip}{3pt}
\setlength{\abovedisplayshortskip}{3pt}
\setlength{\belowdisplayskip}{3pt}
\setlength{\belowdisplayshortskip}{3pt}
\small
    \text{S}(w_i) = P(y_\mathrm{t} \mid \bm s) - P(y_\mathrm{t} \mid \bm s'_{i-1}w_i),
\end{equation}
where  $y_\mathrm{t}$ is the target classification
label of the input $\bm s$ and $P(y_\mathrm{t} \mid *)$ is the classification confidence computed by a PLM fine-tuned on the task. Note that the PLM used here is the model with GenePT introduced in Section \ref{sec:tf}, not a fully pre-trained PLM. 
In experiments, we set $\delta=0.05$. The important token criterion $\text{S}(w_i) < \delta$ means after adding $w_i$, the fine-tuned PLM can correctly classify the incomplete sequence buffer with a close confidence to the complete sequence. 

\subsubsection*{Masking on in-domain unsupervised data}\label{in-domain}
For $D_{\mathrm{Domain}}$, text classification labels needed for computing $P(y_{\mathrm{t}} \mid *)$ are unavailable to perform the method stated above.

To find and mask important tokens of $D_{\mathrm{Domain}}$, we apply the above method to $D_{\mathrm{Task}}$ to generate a small scale of data where important tokens are annotated. Then we fine-tune a PLM on the annotated data to learn the implicit rules for selecting the important tokens of $D_{\mathrm{Domain}}$. The PLM used here is also the model with GenePT. The fine-tuning task here is a binary classification to classify whether a token is importent or not. With this fine-tuned PLM as a scoring function, we can efficiently score each token of $D_{\mathrm{Domain}}$ without labels and select the important tokens to be masked. After masking the important tokens, $D_{\mathrm{Domain}}$ can be used as the training corpus for our task-guided pre-training. 

\begin{figure*}[t]
\centering
\subfigure[Sem14-Rest + Yelp]{
\begin{minipage}[t]{0.25\linewidth}
\centering
\includegraphics[scale=0.16]{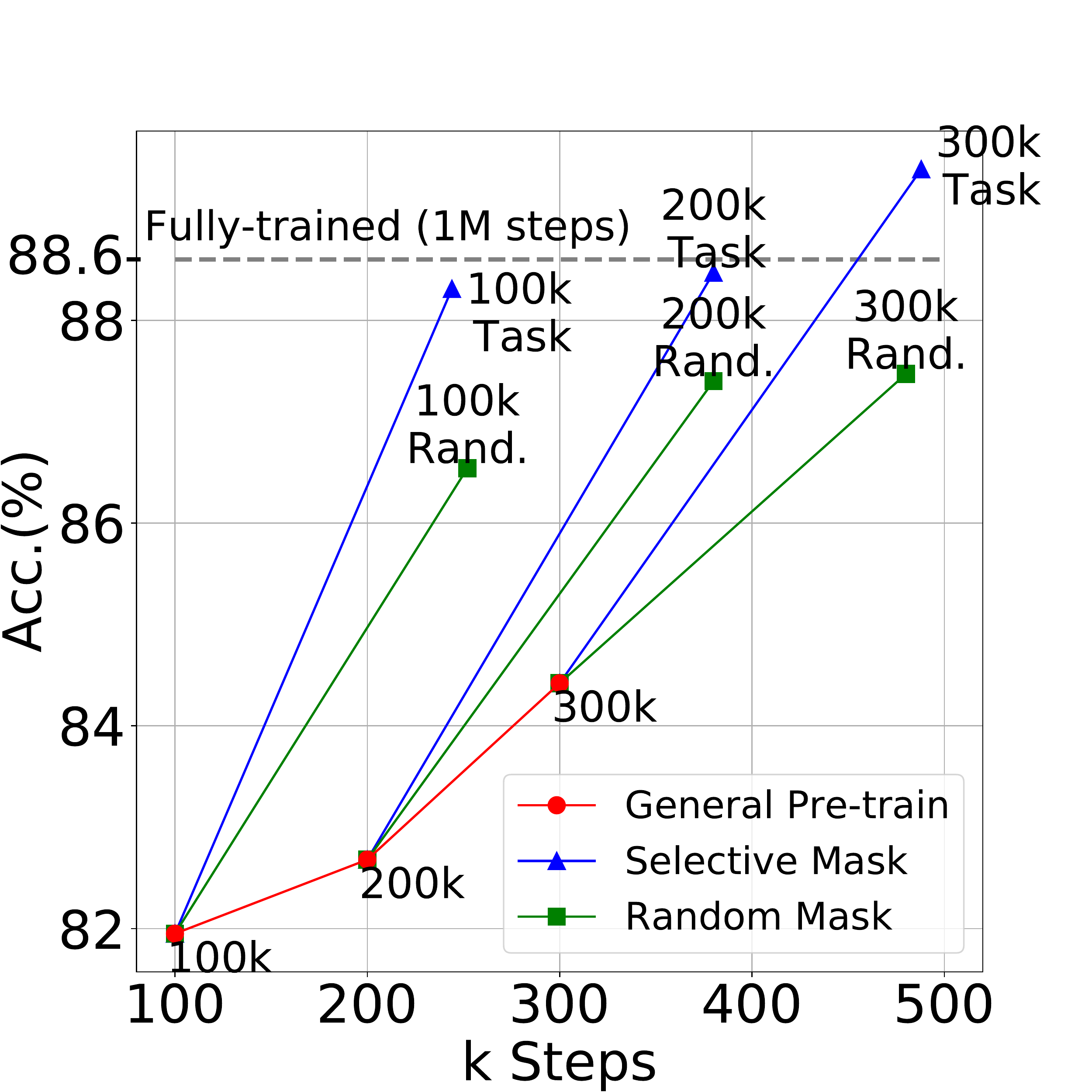}
\end{minipage}%
}%
\subfigure[Sem14-Rest + Amazon]{
\begin{minipage}[t]{0.25\linewidth}
\centering
\includegraphics[scale=0.16]{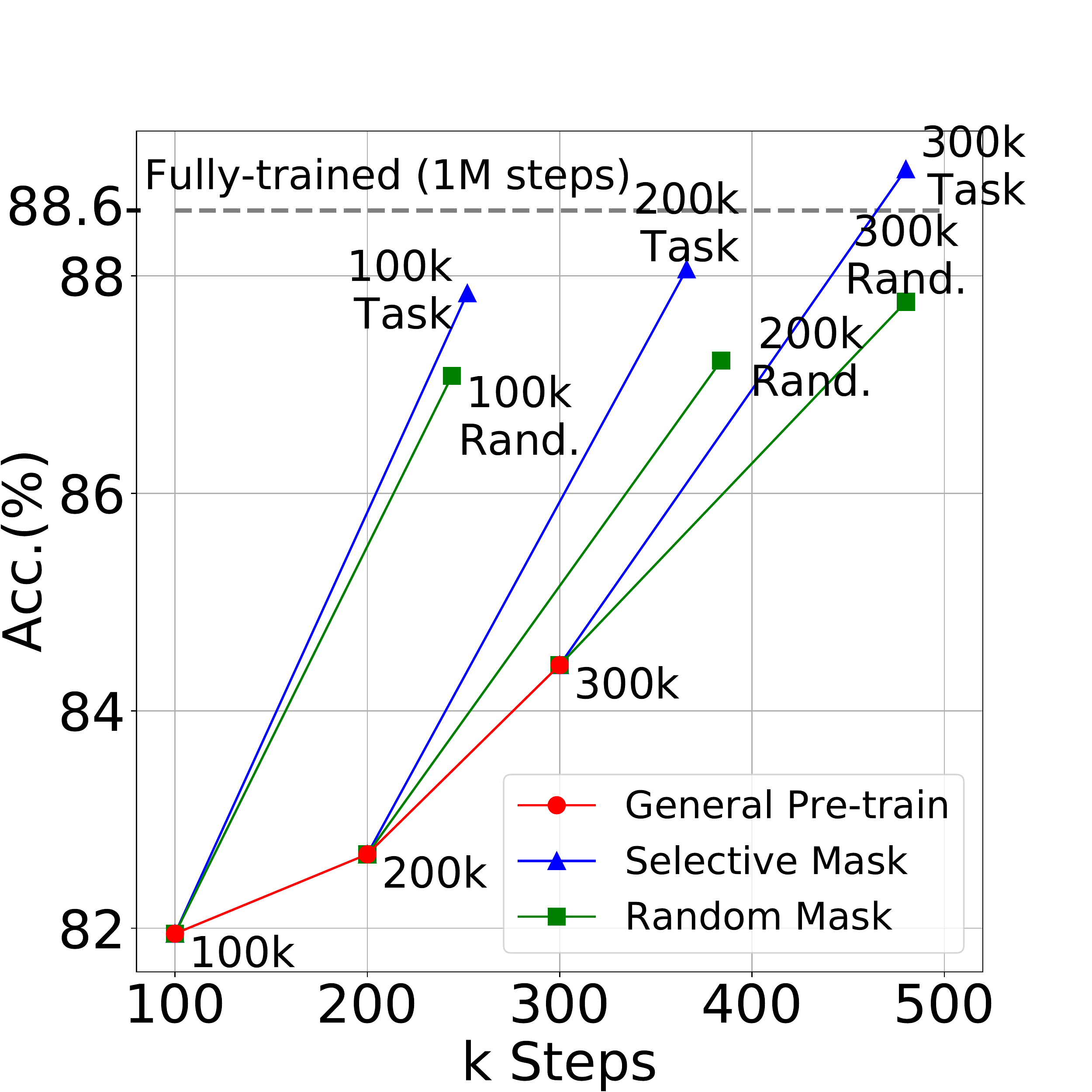}
\end{minipage}%
}%
\subfigure[MR + Yelp]{
\begin{minipage}[t]{0.25\linewidth}
\centering
\includegraphics[scale=0.16]{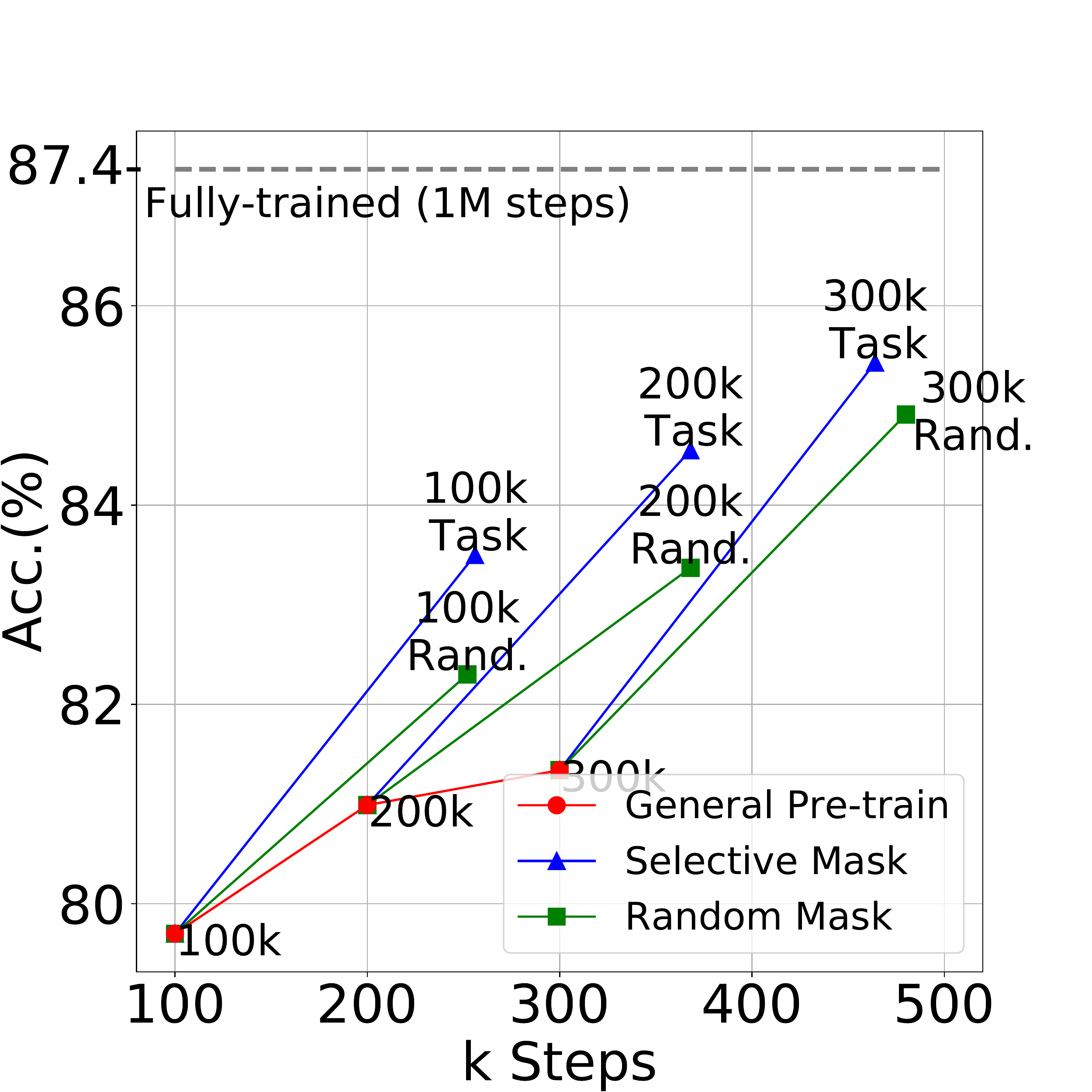}
\end{minipage}
}%
\subfigure[MR + Amazon]{
\begin{minipage}[t]{0.25\linewidth}
\centering
\includegraphics[scale=0.16]{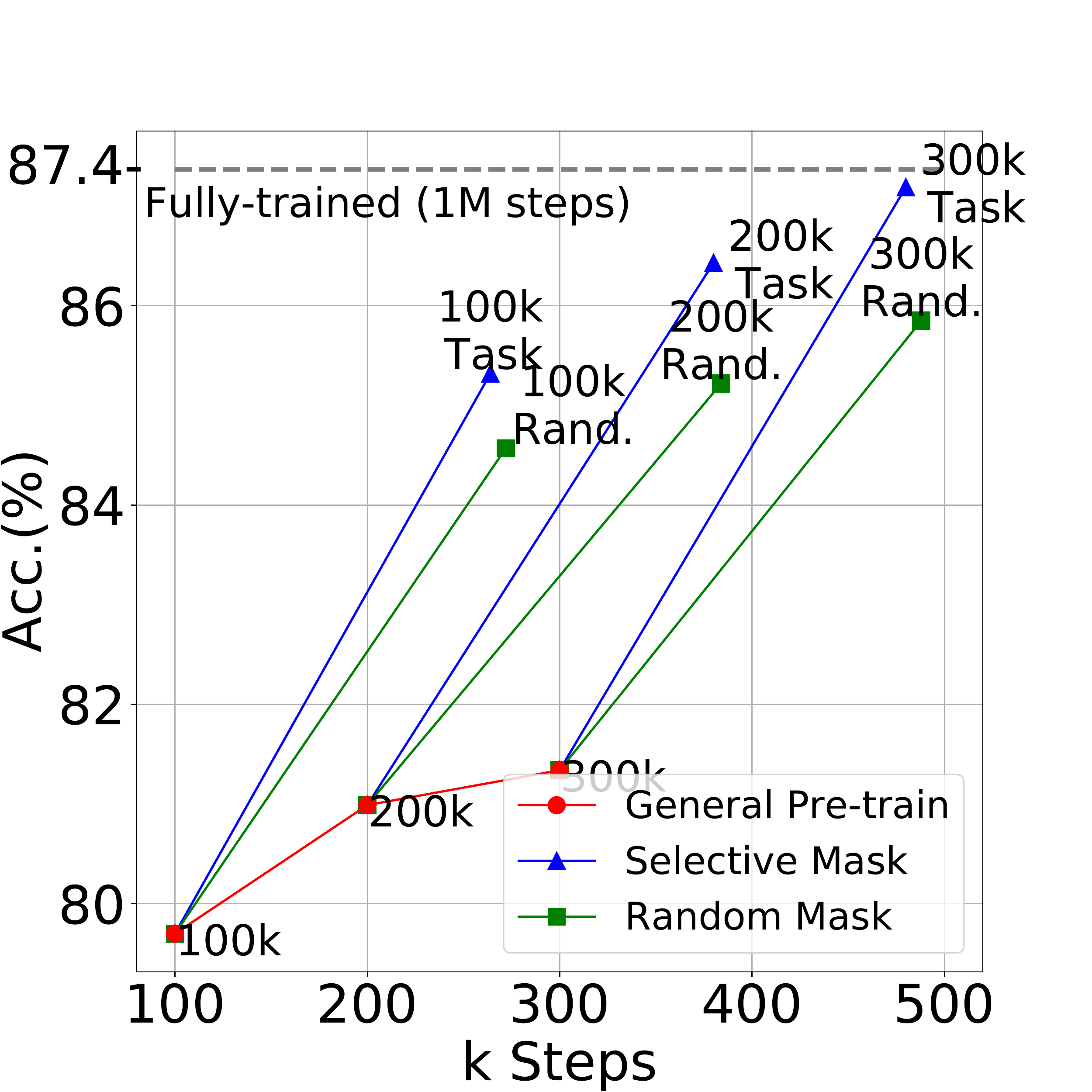}
\end{minipage}
}%
\centering
\caption{Experimental results on 4 different combinations (Task + $D_{\mathrm{Domain}}$). The y-axis indicates the test accuracy. The x-axis indicates the overall pre-training steps. The general pre-training starts at 0 steps and stops at 100k, 200k and 300k steps, corresponding to the ``General Pre-train'' line. Then task-guided pre-training or random mask pre-training runs for about 200k steps, corresponding to the ``Selective Mask'' line and ``Random Mask'' line.  }\label{fig:res}
\end{figure*}

\section{Experiments}
\subsection{Experimental Settings}
\label{sec:es}
We evaluate our method on two sentiment analysis tasks: MR \cite{pang2005seeing} and SemEval14 task 4 restaurant and laptop datasets~\cite{pontiki-etal-2014-semeval}, using the model architecture of $\mathrm{BERT}_\mathrm{BASE}$ in~\citet{devlin2018bert}. Considering the space limit, we only report the results on SemEval14-Restaurant in the main paper. The results on SemEval14-Laptop can be found in the appendix. For simplicity, we abbreviate SemEval14-Restaurant to Sem14-Rest in the rest of the paper. 

In GenePT, we adopt BookCorpus~\cite{Zhu_2015} and English Wikipedia as our $D_{\mathrm{General}}$. To show that our strategy can significantly reduce the computation cost of pre-training, we choose the model which early stopped at 100k, 200k, and 300k steps and the fully pre-trained model (1M steps).

In TaskPT, we use the pure text of Yelp \cite{NIPS2015_5782} and Amazon \cite{he2016ups} as our in-domain unsupervised data $D_{\mathrm{Domain}}$. These two datasets are both 1000 times larger than MR \& Sem14-Rest ($D_{\mathrm{Task}}$), and 100 times smaller than BookCorpus \& English Wikipedia ($D_{\mathrm{General}}$).  

In fine-tuning, we fine-tune the model for 10 epochs and choose the version with the highest accuracy on the dev set. 

\subsection{Experimental Results}
\subsubsection*{Efficiency}
We report our accuracy-pre-training-step lines of all four combinations of downstream tasks and $D_{\mathrm{Domain}}$ in Figure~\ref{fig:res}. Note that since the cost of the selective masking is insignificant compared with that of the task-specific pre-training, it is ignored in Figure~\ref{fig:res}. The detailed analysis of the computation cost in selective masking can be found in the Appendix. From the experimental results, we can observe that 

(1) Our method achieves comparable or even better performances on all 4 settings with less than 50\% pre-training costs, which indicates our task-guided pre-training method with selective masking is both efficient and effective.

(2) Our selective masking strategy consistently outperforms the random selecting strategy that most previous works use, which indicates that our selective masking works well for capturing task-specific language patterns.

(3) In the 4 settings, our model performs best in Sem14-Rest + Yelp, in which our model outperforms the fully pre-trained $\mathrm{BERT}_\mathrm{BASE}$ by 1.4\% with only half of the training steps. While in MR+Yelp, the model performs worst, in which our accuracy drops 1.94\% compared with the fully pre-trained model. This is because the text domains of Sem14-Rest and Yelp are much more similar (both restaurant reviews) than those of MR and Yelp (movie reviews and restaurant reviews). It indicates that the similarity between $D_{\mathrm{Domain}}$ and $D_{\mathrm{Task}}$ is critical for our task-guided pre-training to capture the domain-specific and task-specific patterns, which is intuitive.

\subsubsection*{Effectiveness}
\begin{table}[t]
    \small
    \centering
    \begin{tabular}{ccll}
    \toprule
      &   &  MR & Sem14-Rest\\
    \midrule
    \multicolumn{2}{c}{w/o Task-guided pre-training} &             87.37   & 88.60\\
    \midrule
    \midrule
    \multirow{2}{*}{Amazon} & Random &       88.35   & 90.40\\
                            & Selective &  \textbf{89.51**}   & \textbf{91.56**}\\
    \midrule
    \multirow{2}{*}{Yelp} & Random  &        87.20   &  90.70 \\
                          & Selective  &   \textbf{88.15**}   &   \textbf{91.87*} \\
    \bottomrule
    \end{tabular}
    \caption{Test accuracies of models trained with different methods (without task-guided pre-training or task-guided pre-training with different masking strategies) after full general pre-training (1M steps). $*$ and $**$ indicate statistically significant ($p<.05$ and $p<.001$).}
    \label{tab:res2}
\end{table}

To evaluate the effectiveness of task-guided pre-training, we continue to pre-train the fully pre-trained $\mathrm{BERT}_\mathrm{BASE}$ on the in-domain data. We use the official model in \cite{devlin2018bert} as the fully pre-trained $\mathrm{BERT}_\mathrm{BASE}$. From Table~\ref{tab:res2}, we have the following observations:

(1) Compared with the fully trained GenePT, our model with TaskPT achieves significant improvements in 3 settings no matter which kind of masking strategies is used, which shows the task-guided pre-training can help the model to capture the domain-specific and task-specific patterns. In the MR+Yelp setting, the random masking harms the model performance, which indicates not all patterns in $D_{\mathrm{Domain}}$ can benefit downstream tasks. 


(2) In all settings, our selective masking strategy consistently outperforms the random masking strategy, even on the setting of MR+Yelp. It indicates that our selective masking strategy can still effectively capture helpful task-specific patterns even when the $D_{\mathrm{Domain}}$ is not so close to the $D_{\mathrm{Task}}$.

\subsection{Case Study}
\begin{table}[t]
    \small
    \centering
    \begin{tabular}{c}
    \toprule
            \makecell[l]{Downstream Dataset: MR \\} \\
    \midrule
        \makecell[l]{Text: Constently \textbf{touching}, surprisingly \\ \textbf{funny}, semi-surreal \#\#ist exploration of the \\ \textbf{creative} act. \\} \\
    \midrule
        \makecell[l]{In-domain Dataset: Yelp \\} \\
    \midrule
        \makecell[l]{Text: \textbf{Nice}, clean, simple setup. Limited seating. \\ Cakes are aw \#\#sum! Very \textbf{fresh}. Even \\ have egg \#\#less cakes. Food is \textbf{good} as well. \\ Really like the pan \#\#ner pan \#\#ini. Also \\ other items are \textbf{worth} checking out. \\} \\
    \bottomrule
    \end{tabular}
    \caption{The former one is a sequence masked by our selective method on downstream data. The latter one is a sequence masked by the PLM scoring function. The bold tokens are selected to be masked.}
    \label{tab:case}
\end{table}

To analyze whether our selective masking strategy can successfully find important tokens, we conduct a case study, as shown in Table \ref{tab:case}. In this case, we use MR as the supervised data and Yelp as the unsupervised in-domain data. It shows that our selective masking strategy successfully selects sentiment tokens, which are important for this task, on both supervised and unsupervised data.

\section{Conclusion}

In this paper, we design task-guided pre-training with selective masking and present a three-stage training framework for PLMs. With task-guided pre-training, models can effectively and efficiently learn domain-specific and task-specific patterns, which benefits downstream tasks. Experimental results show that our methods can achieve better performances with less computation cost. Note that although we only conduct experiments on two sentiment classification tasks using BERT as the base model, our method can easily generalize to other models using masked language modeling or its variants and other text classification tasks.

Besides, there are still two important directions for future work: (1)  How to apply task-guided pre-training to general domain data when the in-domain data is limited. 
(2) How to design more effective strategies to capture domain-specific and task-specific patterns for selective masking.

\section*{Acknowledgement}
This work is supported by the National Key Research and Development Program of China (No. 2018YFB1004503), the National Natural Science Foundation of China (NSFC No. 61732008) and Beijing Academy  of Artificial Intelligence (BAAI).

\bibliography{emnlp2020}

\begin{thebibliography}{26}
\expandafter\ifx\csname natexlab\endcsname\relax\def\natexlab#1{#1}\fi

\bibitem[{Aharoni and Goldberg(2020)}]{aharoni2020unsupervised}
Roee Aharoni and Yoav Goldberg. 2020.
\newblock \href {https://arxiv.org/pdf/2004.02105} {Unsupervised domain
  clusters in pretrained language models}.
\newblock In \emph{Proceedings of ACL}.

\bibitem[{Alsentzer et~al.(2019)Alsentzer, Murphy, Boag, Weng, Jindi, Naumann,
  and McDermott}]{alsentzer-etal-2019-publicly}
Emily Alsentzer, John Murphy, William Boag, Wei-Hung Weng, Di~Jindi, Tristan
  Naumann, and Matthew McDermott. 2019.
\newblock \href {https://www.aclweb.org/anthology/W19-1909} {Publicly available
  clinical {BERT} embeddings}.
\newblock In \emph{Proceedings of ClinicalNLP}.

\bibitem[{Baevski et~al.(2019)Baevski, Edunov, Liu, Zettlemoyer, and
  Auli}]{baevski2019cloze}
Alexei Baevski, Sergey Edunov, Yinhan Liu, Luke Zettlemoyer, and Michael Auli.
  2019.
\newblock \href {https://www.aclweb.org/anthology/D19-1539} {Cloze-driven
  pretraining of self-attention networks}.
\newblock In \emph{Proceedings of EMNLP}.

\bibitem[{Beltagy et~al.(2019)Beltagy, Lo, and Cohan}]{Beltagy2019SciBERT}
Iz~Beltagy, Kyle Lo, and Arman Cohan. 2019.
\newblock \href {https://www.aclweb.org/anthology/D19-1371} {{SciBERT}:
  Pretrained language model for scientific text}.
\newblock In \emph{Proceedings of EMNLP}.

\bibitem[{Clark et~al.(2020)Clark, Luong, Le, and Manning}]{Clark2020ELECTRA:}
Kevin Clark, Minh-Thang Luong, Quoc~V. Le, and Christopher~D. Manning. 2020.
\newblock \href {https://openreview.net/pdf?id=r1xMH1BtvB} {Electra:
  Pre-training text encoders as discriminators rather than generators}.
\newblock In \emph{Proceedings of ICLR}.

\bibitem[{Devlin et~al.(2019)Devlin, Chang, Lee, and
  Toutanova}]{devlin2018bert}
Jacob Devlin, Ming-Wei Chang, Kenton Lee, and Kristina Toutanova. 2019.
\newblock \href {https://www.aclweb.org/anthology/N19-1423} {{BERT}:
  Pre-training of deep bidirectional transformers for language understanding}.
\newblock In \emph{Proceedings of NAACL}.

\bibitem[{Feng et~al.(2018)Feng, Wallace, Grissom~II, Iyyer, Rodriguez, and
  Boyd-Graber}]{feng-etal-2018-pathologies}
Shi Feng, Eric Wallace, Alvin Grissom~II, Mohit Iyyer, Pedro Rodriguez, and
  Jordan Boyd-Graber. 2018.
\newblock \href {https://www.aclweb.org/anthology/D18-1407/} {Pathologies of
  neural models make interpretations difficult}.
\newblock In \emph{Proceedings of EMNLP}.

\bibitem[{Gururangan et~al.(2020)Gururangan, Marasović, Swayamdipta, Lo,
  Beltagy, Downey, and Smith}]{gururangan2020dont}
Suchin Gururangan, Ana Marasović, Swabha Swayamdipta, Kyle Lo, Iz~Beltagy,
  Doug Downey, and Noah~A. Smith. 2020.
\newblock \href {https://arxiv.org/pdf/2004.10964} {Don't stop pretraining:
  Adapt language models to domains and tasks}.
\newblock In \emph{Proceedings of ACL}.

\bibitem[{He and McAuley(2016)}]{he2016ups}
Ruining He and Julian McAuley. 2016.
\newblock \href {https://dl.acm.org/doi/10.1145/2872427.2883037} {Ups and
  downs: Modeling the visual evolution of fashion trends with one-class
  collaborative filtering}.
\newblock In \emph{Proceedings of WWW}.

\bibitem[{Huang et~al.(2020)Huang, Altosaar, and
  Ranganath}]{huang2019clinicalbert}
Kexin Huang, Jaan Altosaar, and Rajesh Ranganath. 2020.
\newblock \href {https://www.chilconference.org/workshop_ws13.html}
  {{ClinicalBERT}: Modeling clinical notes and predicting hospital
  readmission}.
\newblock In \emph{Proceedings of ACM-CHIL}.

\bibitem[{Joshi et~al.(2020)Joshi, Chen, Liu, Weld, Zettlemoyer, and
  Levy}]{joshi2019spanbert}
Mandar Joshi, Danqi Chen, Yinhan Liu, Daniel~S Weld, Luke Zettlemoyer, and Omer
  Levy. 2020.
\newblock \href {https://www.aclweb.org/anthology/2020.tacl-1.5/} {Span{BERT}:
  Improving pre-training by representing and predicting spans}.
\newblock \emph{Transactions of the Association for Computational Linguistics},
  8:64--77.

\bibitem[{Ke et~al.(2020)Ke, Ji, Liu, Zhu, and Huang}]{ke2019sentilr}
Pei Ke, Haozhe Ji, Siyang Liu, Xiaoyan Zhu, and Minlie Huang. 2020.
\newblock \href {https://arxiv.org/abs/1911.02493} {Senti{LARE}:
  Sentiment-aware language representation learning with linguistic knowledge}.
\newblock In \emph{Proceedings of EMNLP}.

\bibitem[{Lee et~al.(2019)Lee, Yoon, Kim, Kim, Kim, So, and
  Kang}]{Lee2019BioBERT}
Jinhyuk Lee, Wonjin Yoon, Sungdong Kim, Donghyeon Kim, Sunkyu Kim, Chan~Ho So,
  and Jaewoo Kang. 2019.
\newblock \href
  {https://academic.oup.com/bioinformatics/article/36/4/1234/5566506}
  {{{BioBERT}: a pre-trained biomedical language representation model for
  biomedical text mining}}.
\newblock \emph{Bioinformatics}.

\bibitem[{Liu et~al.(2019)Liu, Ott, Goyal, Du, Joshi, Chen, Levy, Lewis,
  Zettlemoyer, and Stoyanov}]{liu2019roberta}
Yinhan Liu, Myle Ott, Naman Goyal, Jingfei Du, Mandar Joshi, Danqi Chen, Omer
  Levy, Mike Lewis, Luke Zettlemoyer, and Veselin Stoyanov. 2019.
\newblock \href {http://arxiv.org/pdf/1907.11692.pdf} {Ro{BERT}a: A robustly
  optimized bert pretraining approach}.
\newblock \emph{arXiv preprint arXiv:1907.11692}.

\bibitem[{Pang and Lee(2005)}]{pang2005seeing}
Bo~Pang and Lillian Lee. 2005.
\newblock \href {http://www.cs.cornell.edu/home/llee/papers/pang-lee-stars.pdf}
  {Seeing stars: Exploiting class relationships for sentiment categorization
  with respect to rating scales}.
\newblock In \emph{Proceedings of ACL}.

\bibitem[{Phang et~al.(2018)Phang, F{\'e}vry, and Bowman}]{phang2018sentence}
Jason Phang, Thibault F{\'e}vry, and Samuel~R Bowman. 2018.
\newblock \href {http://arxiv.org/pdf/1811.01088.pdf} {Sentence encoders on
  stilts: Supplementary training on intermediate labeled-data tasks}.
\newblock \emph{arXiv preprint arXiv:1811.01088}.

\bibitem[{Pontiki et~al.(2014)Pontiki, Galanis, Pavlopoulos, Papageorgiou,
  Androutsopoulos, and Manandhar}]{pontiki-etal-2014-semeval}
Maria Pontiki, Dimitris Galanis, John Pavlopoulos, Harris Papageorgiou, Ion
  Androutsopoulos, and Suresh Manandhar. 2014.
\newblock \href {https://www.aclweb.org/anthology/S14-2004} {{S}em{E}val-2014
  task 4: Aspect based sentiment analysis}.
\newblock In \emph{Proceedings of SemEval14}.

\bibitem[{Radford et~al.(2018)Radford, Narasimhan, Salimans, and
  Sutskever}]{radford2018improving}
Alec Radford, Karthik Narasimhan, Tim Salimans, and Ilya Sutskever. 2018.
\newblock \href
  {https://s3-us-west-2.amazonaws.com/openai-assets/research-covers/language-unsupervised/language_understanding_paper.pdf}
  {Improving language understanding by generative pre-training}.
\newblock In \emph{Proceedings of Technical report, OpenAI}.

\bibitem[{Rietzler et~al.(2020)Rietzler, Stabinger, Opitz, and
  Engl}]{rietzler2020adapt}
Alexander Rietzler, Sebastian Stabinger, Paul Opitz, and Stefan Engl. 2020.
\newblock \href
  {http://www.lrec-conf.org/proceedings/lrec2020/pdf/2020.lrec-1.607.pdf}
  {Adapt or get left behind: Domain adaptation through bert language model
  finetuning for aspect-target sentiment classification}.
\newblock In \emph{Proceedings of LREC}.

\bibitem[{Sung et~al.(2019)Sung, Dhamecha, Saha, Ma, Reddy, and
  Arora}]{sung-etal-2019-pre}
Chul Sung, Tejas Dhamecha, Swarnadeep Saha, Tengfei Ma, Vinay Reddy, and Rishi
  Arora. 2019.
\newblock \href {https://www.aclweb.org/anthology/D19-1628} {Pre-training
  {BERT} on domain resources for short answer grading}.
\newblock In \emph{Proceedings of EMNLP}.

\bibitem[{Wang et~al.(2019)Wang, Hula, Xia, Pappagari, McCoy, Patel, Kim,
  Tenney, Huang, Yu, Jin, Chen, Van~Durme, Grave, Pavlick, and
  Bowman}]{wang2019can}
Alex Wang, Jan Hula, Patrick Xia, Raghavendra Pappagari, R.~Thomas McCoy, Roma
  Patel, Najoung Kim, Ian Tenney, Yinghui Huang, Katherin Yu, Shuning Jin,
  Berlin Chen, Benjamin Van~Durme, Edouard Grave, Ellie Pavlick, and Samuel~R.
  Bowman. 2019.
\newblock \href {https://www.aclweb.org/anthology/P19-1439} {Can you tell me
  how to get past sesame street? {S}entence-level pretraining beyond language
  modeling}.
\newblock In \emph{Proceedings of ACL}.

\bibitem[{Xu et~al.(2019)Xu, Liu, Li, Poon, and
  Gao}]{xu-etal-2019-doubletransfer}
Yichong Xu, Xiaodong Liu, Chunyuan Li, Hoifung Poon, and Jianfeng Gao. 2019.
\newblock \href {https://www.aclweb.org/anthology/W19-5042} {{D}ouble{T}ransfer
  at {MEDIQA} 2019: Multi-source transfer learning for natural language
  understanding in the medical domain}.
\newblock In \emph{Proceedings of BioNLP}.

\bibitem[{Yang et~al.(2019)Yang, Dai, Yang, Carbonell, Salakhutdinov, and
  Le}]{yang2019xlnet}
Zhilin Yang, Zihang Dai, Yiming Yang, Jaime Carbonell, Ruslan Salakhutdinov,
  and Quoc~V Le. 2019.
\newblock \href
  {http://papers.nips.cc/paper/8812-xlnet-generalized-autoregressive-pretraining-for-language-understanding.pdf}
  {{XLNet}: Generalized autoregressive pretraining for language understanding}.
\newblock In \emph{Proceedings of NeurlPS}.

\bibitem[{Zhang et~al.(2015)Zhang, Zhao, and LeCun}]{NIPS2015_5782}
Xiang Zhang, Junbo Zhao, and Yann LeCun. 2015.
\newblock \href
  {http://papers.nips.cc/paper/5782-character-level-convolutional-networks-for-text-classification}
  {Character-level convolutional networks for text classification}.
\newblock In \emph{Proceedings of NeurlPS}.

\bibitem[{Zhu et~al.(2015)Zhu, Kiros, Zemel, Salakhutdinov, Urtasun, Torralba,
  and Fidler}]{Zhu_2015}
Yukun Zhu, Ryan Kiros, Rich Zemel, Ruslan Salakhutdinov, Raquel Urtasun,
  Antonio Torralba, and Sanja Fidler. 2015.
\newblock \href
  {https://www.computer.org/csdl/proceedings-article/iccv/2015/8391a019/12OmNro0HYa}
  {Aligning books and movies: Towards story-like visual explanations by
  watching movies and reading books}.
\newblock In \emph{Proceedings of ICCV}.

\bibitem[{Ziser and Reichart(2018)}]{ziser-reichart-2018-pivot}
Yftah Ziser and Roi Reichart. 2018.
\newblock \href {https://www.aclweb.org/anthology/N18-1112} {Pivot based
  language modeling for improved neural domain adaptation}.
\newblock In \emph{Proceedings of NAACL}.

\end{thebibliography}
\bibliographystyle{acl_natbib}

\appendix

\section*{Appendices}
\section{Cost of Selective Masking}
In practice, our selective masking method described in Section \ref{sec:sm} can be implemented in the following 4 steps:
\begin{itemize}
    \item \textbf{Fine-tune BERT}. Fine-tune a checkpoint after the GenePT on downstream supervised datasets (i.e., MR \& SemEval14).
    \item \textbf{Downstream Mask}. Selectively annotate important tokens on downstream supervised datasets using the method stated in ``Finding important tokens". 
    \item \textbf{Train NN}. Train a token-level binary classification BERT model from the checkpoint after the GenePT on downstream supervised datasets where important tokens are annotated.
    \item \textbf{In-domain Mask}. Use the token-level binary classification model trained in the \textbf{Train NN} step to select important tokens on in-domain unsupervised datasets (i.e., Yelp \& Amazon), and mask them.
\end{itemize}

The additional time cost introduced by the 4 steps in selective masking strategy is shown in Table \ref{tab:time}. From the table, we conclude that the extra computation time cost of our selective masking strategy is insignificant compared with the cost saved in the pre-training stage, so we ignore it in the calculation and comparison of pre-training steps.

\begin{table}[h]
    \small
    \centering
    \begin{tabular}{ccccc}
    \toprule
     & \multicolumn{2}{c}{MR} & \multicolumn{2}{c}{SemEval14} \\
     & Yelp & Amazon & Yelp & Amazon\\
    \midrule
    Finetune BERT & 10  & 10 & 3 & 3\\
    Downstream Mask & 20  &  20  & 10 & 10\\
    Train NN & 10  &  10 & 3 & 3\\
    In-domain Mask & 40   & 120 & 40 & 120\\
    \midrule
    Sum & 70  & 150 & 56 & 136\\
    \midrule
    \midrule
    Saved Cost & 2160  & 2160 & 2160 & 2160\\
    \bottomrule
    \end{tabular}
    \caption{The comparison between the cost of the 4 steps for tokens selection and that saved by our selective masking method (in minutes). The 1-5 lines are the time for every stage and their summation of token selection. The last line is the saved pre-training time.}
    \label{tab:time}
\end{table}

\begin{figure*}[t]
\centering
\subfigure[SemEval14 + Yelp]{
\begin{minipage}[t]{0.5\linewidth}
\centering
\includegraphics[scale=0.25]{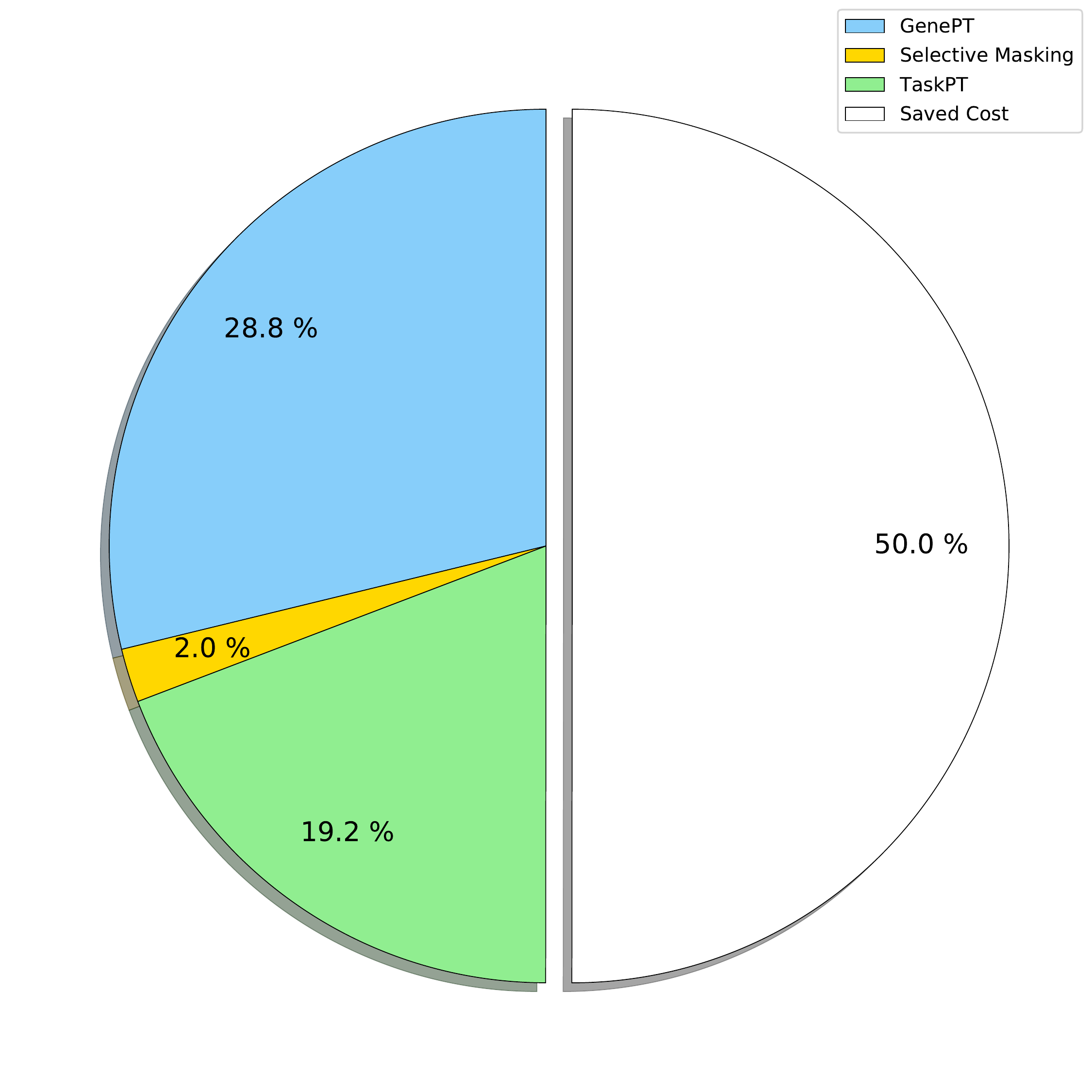}
\end{minipage}%
}%
\subfigure[SemEval14 + Amazon]{
\begin{minipage}[t]{0.5\linewidth}
\centering
\includegraphics[scale=0.25]{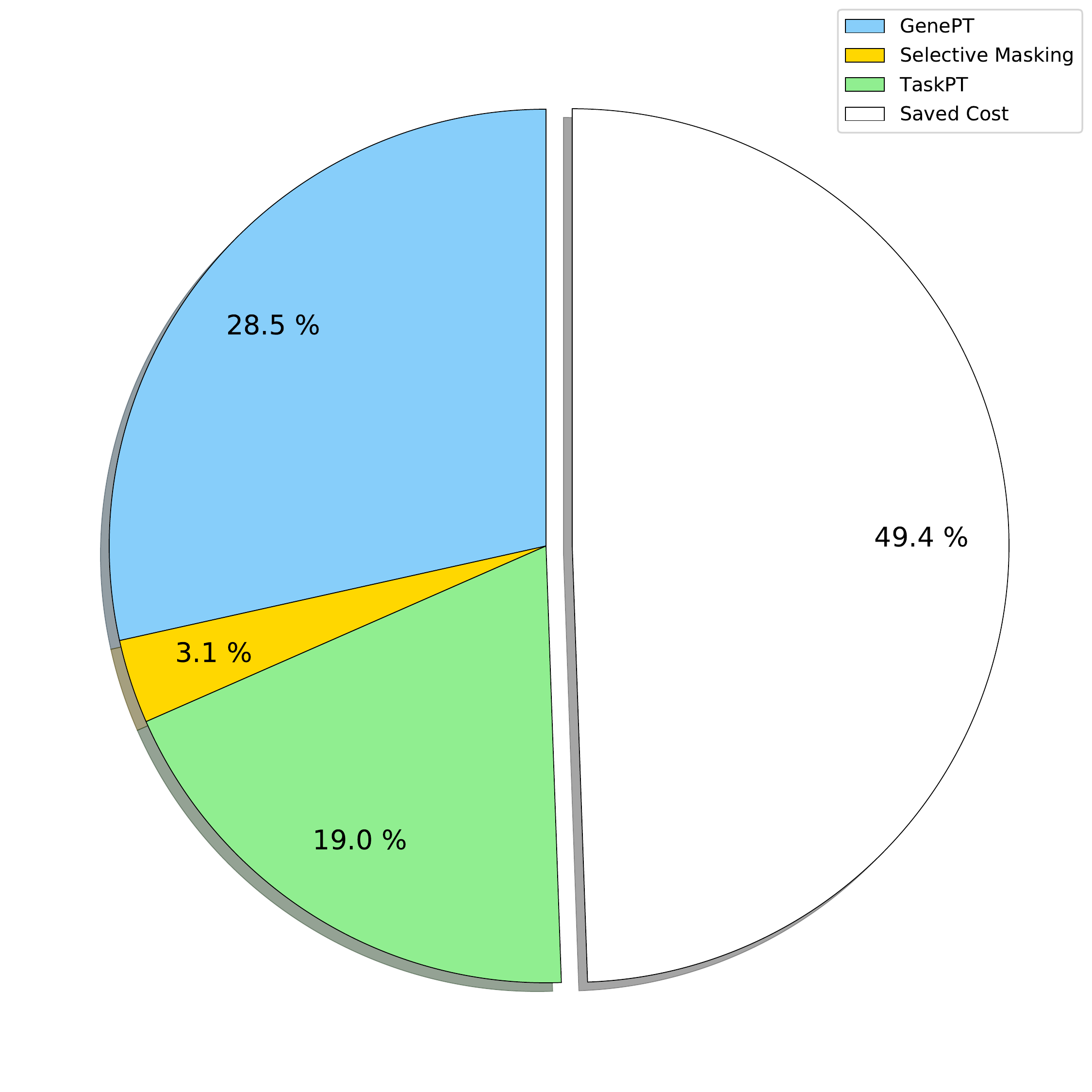}
\end{minipage}%
}%

\subfigure[MR + Yelp]{
\begin{minipage}[t]{0.5\linewidth}
\centering
\includegraphics[scale=0.25]{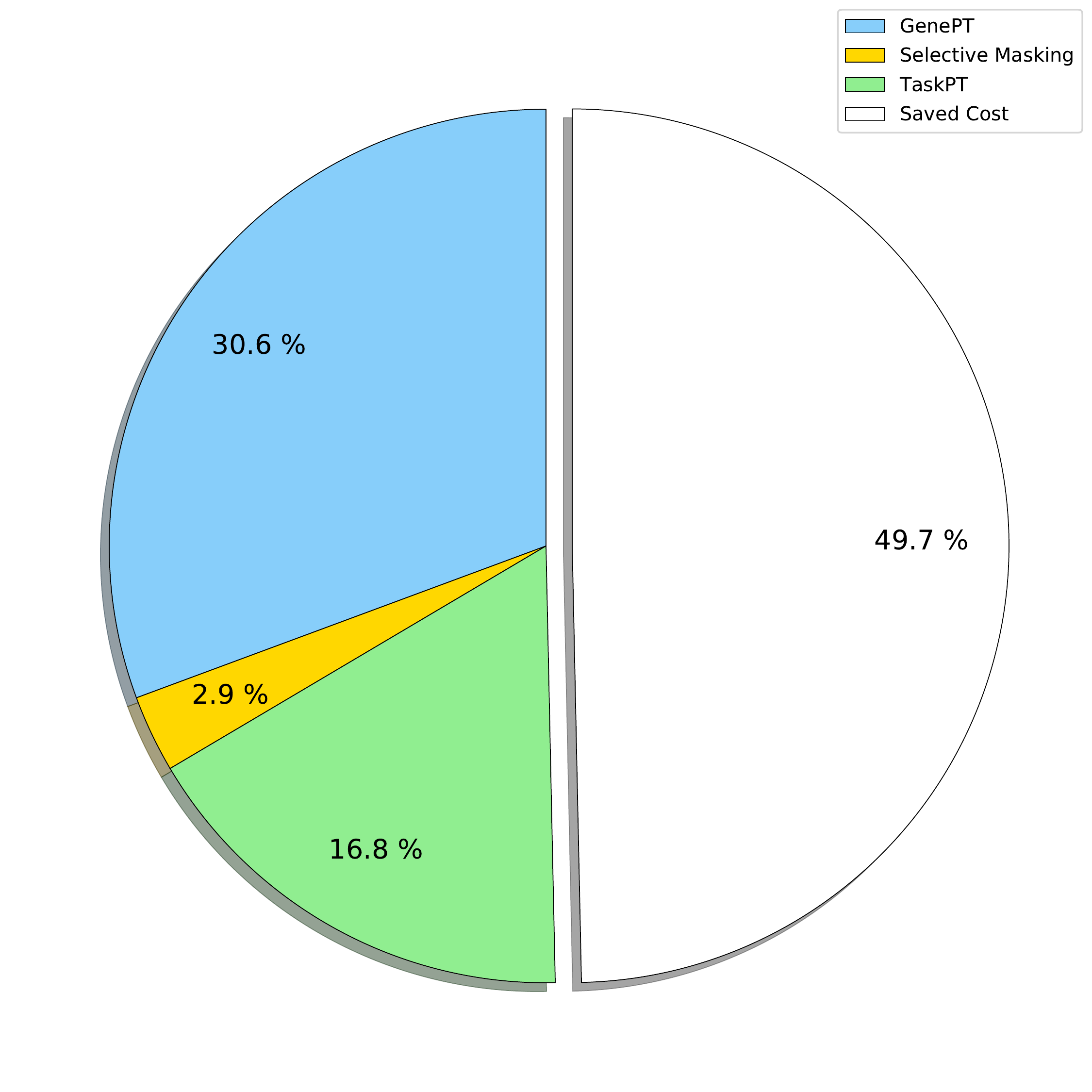}
\end{minipage}
}%
\subfigure[MR + Amazon]{
\begin{minipage}[t]{0.5\linewidth}
\centering
\includegraphics[scale=0.25]{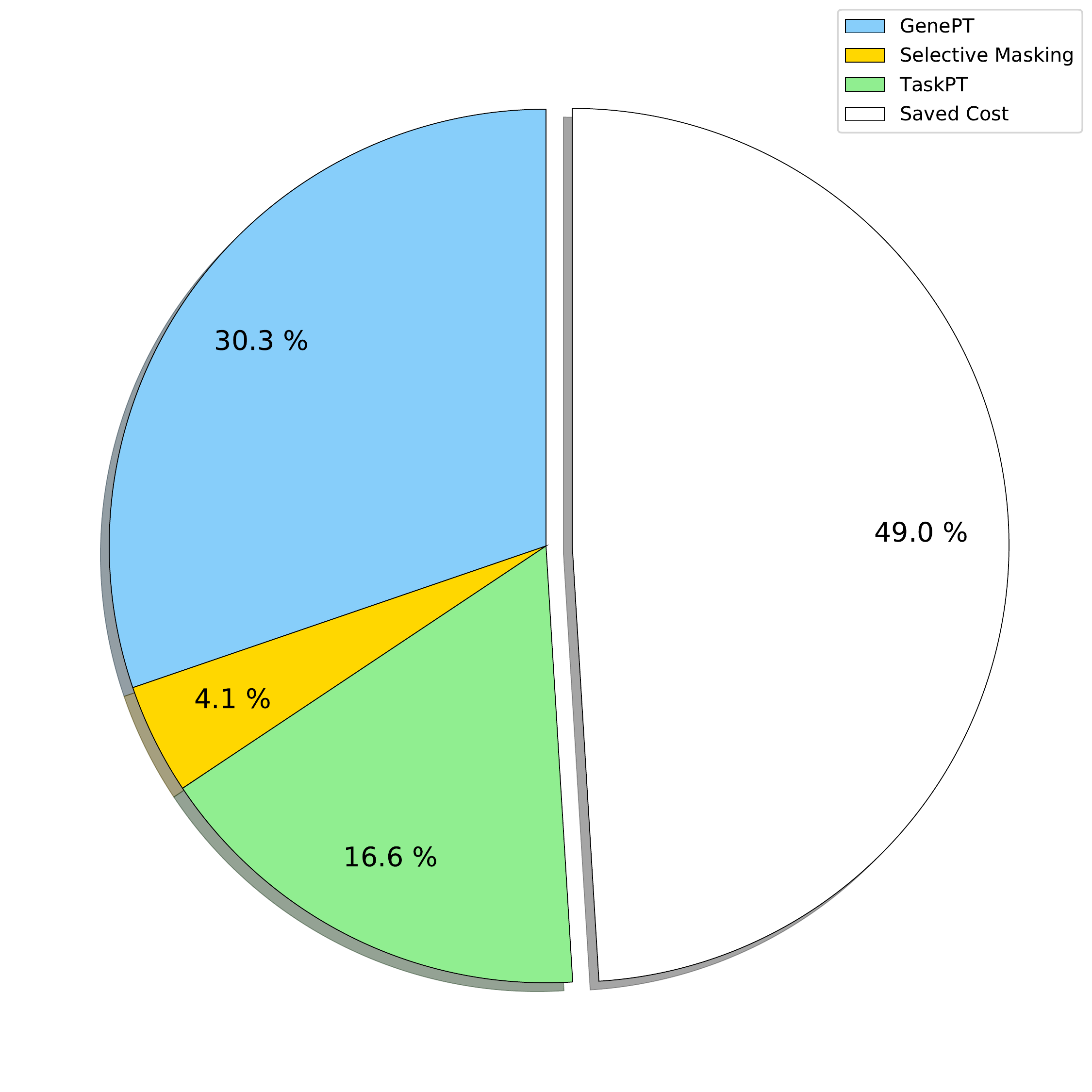}
\end{minipage}
}%
\centering
\caption{The proportion of the time cost of different pre-training stages 4 different combinations (Task + $D_{\mathrm{Domain}}$). The whole pie represents the time cost of the conventional pre-training method. The colored sectors represent the time cost of GenePT, selective masking, and TaskPT respectively. The white sector shows the time saved by our training framework.}\label{fig:time}
\end{figure*}

In Figure \ref{fig:time}, we also illustrate the proportion of different stages according to the time they spend in our experiments. The whole pie is the conventional random-masking pre-training cost and the colored sectors are the time cost of the proposed GenePT, selective masking strategy, and TaskPT. The white sector, as a result, indicates the pre-training time saved in our training framework. From the figures, we can see that the cost of selective masking only contributes a small part of the whole pre-training time and about half of the conventional pre-training cost (about 36 hours in our experiments) is saved with our method.

\section{Detailed Experimental Setup}
\subsection{GenePT}
We generally followed the pre-training procedure and hyper-parameters of $\mathrm{BERT}_\mathrm{BASE}$ in \cite{devlin2018bert} except that we set the max tokens number in a sequence to 256 and utilized FP16 precision\footnote{\url{https://github.com/NVIDIA/DeepLearningExamples/tree/master/PyTorch/LanguageModeling/BERT}} for efficiency. We pre-trained the model on 4 NVIDIA V100 GPUs. The whole 1M-step training completes in about 3 days and we saved checkpoints at 100k, 200k, and 300k steps during the process.

\subsection{Selective Masking.}
The implementation details of each steps in selective masking are described as follows.
\paragraph{Fine-tune BERT.} 
\label{ftb}
We fine-tuned the checkpoint that stopped GenePT at 100k, 200k, 300k, and 1M steps respectively on downstream supervised datasets MR and SemEval14 with the same hyper-parameters. The fine-tuning batch size was 64 with max tokens number 256. The learning rate was 2e-5 and we used 42 as the random seed. We fine-tuned for at most 10 epochs and selected the model with the highest accuracy on valid datasets. 
    

\paragraph{Downstream Mask.}
We used the models after being fine-tuned on MR and SemEval14 as classifiers to perform the important tokens selecting method on downstream supervised datasets. The sentences were tokenized by the BERT's sub-word tokenizer. We set $\delta = 0.05$ in all circumstances. After the selection, important tokens were annotated as label ``1" while others were labeled as ``0". 

\paragraph{Train NN.}
We added a token-level binary classification head on the top of the BERT checkpoints after GenePT and fine-tuned them with the annotated data after the Downstream Mask stage. The max sequence length was 128, with batch size 64 and learning rate 1e-5. Besides, to balance the two labels, we set the weight 1.5 for label ``1"(important tokens).

\paragraph{In-domain Mask.}
The NN-based token selection was performed by classifying each token in mid-scale in-domain datasets with the model trained after the \textbf{Train NN} stage. If the classification result was ``1", then the token was regarded as important and would be masked in the pre-training stage afterward.

\subsection{TaskPT}
We then continued pre-training the checkpoints after GenePT on selectively masked in-domain datasets. The hyper-parameters were almost the same as that in GenePT, except that we only pre-trained for at most 200k steps.

\subsection{Fine-tuning}
The model after TaskPT was then fine-tuned on downstream datasets MR and SemEval14. The hyper-parameters were generally the same with the \textbf{Fine-tune BERT} stage (\ref{ftb}) except that we averaged the model performance over 10 different random seeds: [13, 43, 83, 181, 271, 347, 433, 659, 727, 859] to provide more convincing results.

\section{Detailed Datasets Description}
We utilized 4 sentiment classification datasets in our experiments. The train/dev/test splits and other statistical information of the 4 datasets are shown in Table \ref{tab:data}.

\begin{table}[h]
    \small
    \centering
    \begin{tabular}{ccc}
    \toprule
    Dataset & Amount & Classes \\
    \midrule
      MR   &  8534/1078/1050  & 2\\
    SemEval14  & 3333/185/973 & 3\\
    Yelp & 700k & 5 \\
    Amazon & 3M & 5 \\
    \bottomrule
    \end{tabular}
    \caption{Datasets statistics. Note that we only use the pure text in the training set of Yelp and Amazon as in-domain unsupervised data}
    \label{tab:data}
\end{table}

\paragraph{MR} MR\footnote{\url{http://www.cs.cornell.edu/people/pabo/movie-review-data/}} is movie-review data for the use in sentiment-analysis experiments. Since the originally released data does not provide train/dev/test split, we randomly sampled 80\% of the whole set for training, 10\% for validation, and 10\% for testing. 

\paragraph{SemEval14} SemEval14\footnote{\url{http://alt.qcri.org/semeval2014/task4/index.php?id=data-and-tools}} is the restaurant-domain dataset released by the task 4 in SemEval14 competition. The original task is aspect-based sentiment analysis. To convert it into a conventional sentiment classification task, we concatenated the aspect tokens and text tokens to form a full sentence as the input to the model.

\paragraph{Yelp} Yelp\footnote{\url{https://www.kaggle.com/yelp-dataset/yelp-dataset}} is a 5-class sentiment classification dataset of reviews about restaurants obtained from the Yelp Dataset Challenge in 2015. In our experiments, we only used its pure text as in-domain unsupervised data.

\paragraph{Amazon} Amazon\footnote{\url{http://jmcauley.ucsd.edu/data/amazon/}} is composed of different reviews on the Amazon website. Similar to Yelp, we only used its pure text to construct in-domain unsupervised data.

\begin{figure}[t]
\centering
\subfigure[Sem14-Lap + Yelp]{
\begin{minipage}[t]{0.5\linewidth}
\centering
\includegraphics[scale=0.16]{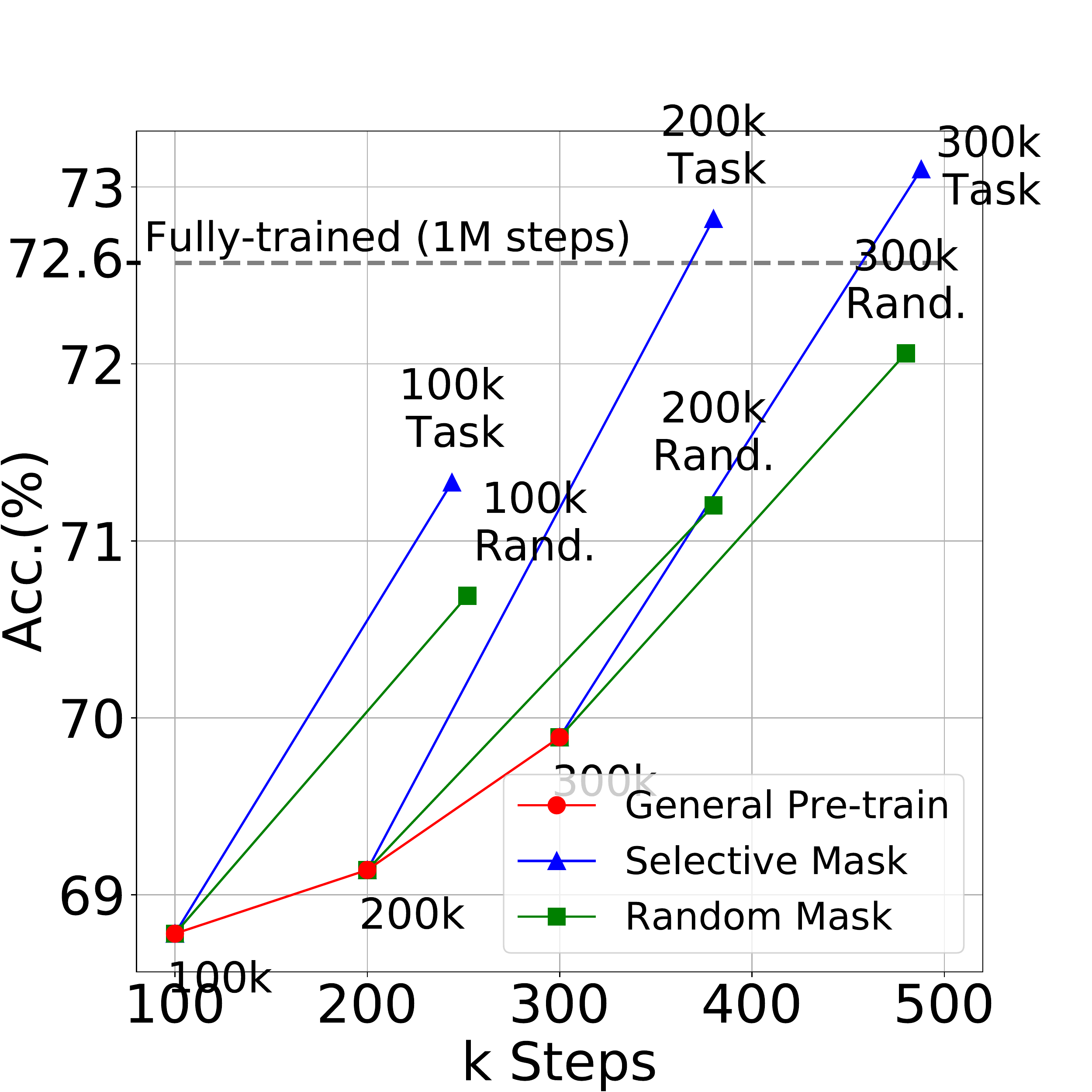}
\end{minipage}%
}%
\subfigure[Sem14-Lap + Amazon]{
\begin{minipage}[t]{0.5\linewidth}
\centering
\includegraphics[scale=0.16]{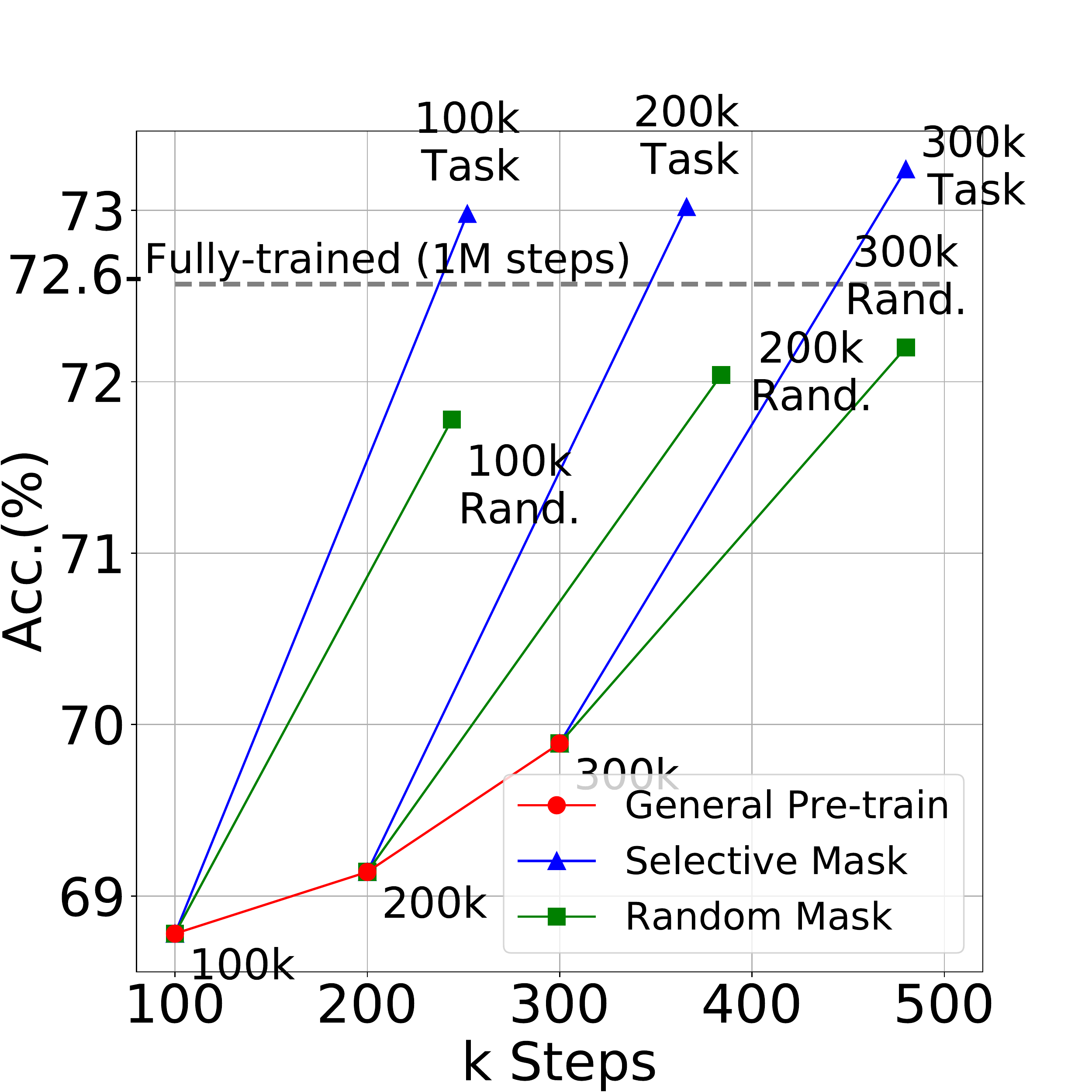}
\end{minipage}%
}%
\centering
\caption{Experimental results on Sem14-Lap + Yelp and Sem14-Lap + Amazon. The y-axis indicates the test accuracy. The x-axis indicates the overall pre-training steps. The general pre-training starts at 0 steps and stops at 100k, 200k and 300k steps, corresponding to the ``General Pre-train'' line. Then task-guided pre-training or random mask pre-training runs for about 200k steps, corresponding to the ``Selective Mask'' line and ``Random Mask'' line. }\label{fig:lap}
\end{figure}

\section{Results on SemEval14-Laptop}

Here we present additional experimental results on the SemEval14 task 4 laptop dataset. We use SemEval14-Laptop as the downstream task dataset and use Yelp and Amazon as in-domain datasets respectively. Similar to Section \ref{sec:es}, we applied our task-guided pre-training to the $\textrm{BERT}_{\textrm{BASE}}$ model early stopped pre-training at 100k, 200k, 300k steps to evaluate the efficiency of our method and also continued to pre-train from the fully pre-trained $\textrm{BERT}_{\textrm{BASE}}$ to evaluate the effectiveness. The accuracy-pre-training-step lines of the efficiency experiment are reported in Figure \ref{fig:lap} and the accuracies in the effectiveness experiment are shown in Table \ref{lap}.

\begin{table}[ht]
    \small
    \centering
    \begin{tabular}{ccc}
    \toprule
                &            & Sem14-Lap \\
    \midrule
     \multicolumn{2}{c}{w/o Task-guided pre-training}     & 72.57 \\
    \midrule
    \midrule
    \multirow{2}{*}{Amazon}  & Random &  73.22 \\
                            & Selctive & \textbf{74.15} \\
    \midrule
    \multirow{2}{*}{Yelp}   & Random    & 73.73 \\
                            & Selective & \textbf{75.26} \\
    \bottomrule
    \end{tabular}
    \caption{Test accuracies of models trained with different
methods (without task-guided pre-training or taskguided
pre-training with different masking strategies) after full general pre-training (1M steps).}
    \label{lap}
\end{table}

From the results, we can conclude that our method is also both effective and efficient on the Sem14-Lap dataset, reaching a better performance with less than 50\% training cost.

\end{document}